# A Virtual Member of a Community of Practice for the Society of Petroleum Engineers: From Prototype to Deployment

**John Boden [1], Joshua Eckroth[1], Dayne Freitag[2], Skyler Gipson[1], Jonathan Keefe[3], Karen Myers[2], Eric Schoen[1], Pedro Sequeira[2], Reid Smith[1], Michael Wessel[2]**

[1]i2k Connect Inc.
[2]SRI International
[3]Pacific Science & Engineering

**Abstract**

We describe the evolution of a virtual assistant, called ATHENA, designed to support the capture, retrieval, and dissemination of knowledge for members of a Community of Practice (CoP) related to the Oil & Gas sector. An evaluation of a first prototype involving 75 professionals from the Society of Petroleum Engineering (SPE) showed that ATHENA dramatically improved both their productivity and performance quality on a set of realistic well-planning tasks compared to their use of a state-of-the-art RAG baseline system. However, the evaluation also identified areas for improvement. This paper describes technical advances to our first prototype in the areas of multi-document retrieval, support for answer validation, and more focused proactive dissemination. Evaluation results show that this enhanced version of ATHENA provides better support for completing knowledge-intensive tasks related to well planning than does a state-of-the-art baseline. ATHENA has been integrated into the SPE Research Portal and is being deployed for use by the society's membership.

## Introduction

We previously developed an intelligent assistant called ATHENA that operates as a *virtual member of a Community of Practice* (CoP) (Wenger 1998; Allee 2000) by modeling how experts naturally seek and share knowledge (Eckroth et al. 2025). ATHENA combines three core capabilities embedded within a chat-based user interface that mirrors the natural workflow of person-to-person knowledge sharing:

- *an agentic approach to information* retrieval that delivers accurate answers rather than raw data, along with rationale and links to source material;
- *in-the-flow capture of experiential knowledge*, or "insights", that can be shared within a CoP;
- *proactive dissemination of insights* tailored to an individual's expertise level and task context via user cognitive modeling and large language models (LLMs).

Our objective was to develop capabilities that could be used broadly. However, to ensure their utility in real-world settings we focused from the start on addressing knowledge management challenges for the Oil & Gas sector with a particular focus on well planning. Well planning involves designing and mapping out the wellbore, estimating costs, and identifying the optimal drilling approach to access hydrocarbons efficiently and safely (Devereux 1998). These tasks require a comprehensive analysis of geological data, regulatory guidelines, and potential environmental impacts. Finding the right information is paramount, as accurate data helps minimize risks, optimize resources, reduce costs, and ensure compliance with safety and environmental standards.

We worked with a corpus of more than 27,000 documents containing general petroleum engineering knowledge drawn from PetroWiki (OnePetro 2025), regulatory matter (US Government 2025), and an extensive data set comprising logs, daily and end-of-well reports, and studies from drilling operations in the Volve North Sea oil field from 2008 to 2016 (Equinor 2025).

Our focus on knowledge management for Oil & Gas was facilitated by members of our team having decades of experience in knowledge management for that sector, including having developed the current research portal for the Society of Petroleum Engineers (SPE), which is the premiere professional society devoted to petroleum engineering with 132,000 members across 146 countries.

Leveraging those connections, we evaluated our initial ATHENA prototype using 75 volunteers from the SPE engaged in realistic well-planning tasks. The results showed that ATHENA significantly improved outcomes (152% increase in task scores) for experts in the domain compared to their use of a then state-of-the-art RAG (Lewis et al. 2020) baseline built on Claude Haiku 3.0. Our technology also increased subject productivity by 283%. Surprisingly, a supplementary evaluation involving 8 subjects showed that ATHENA enabled non-experts to complete tasks as quickly and accurately as experts. The basis for these results was a combination of (1) an agentic chat capability that combined a RAG architecture with faceted search (Hearst 2009), providing much more accurate answers than the baseline, and (2) timely dissemination of insights that provided meaningful guidance tailored to the problem-solving context. Most importantly, 79% of the SPE participants recommended in favor of adoption of the technology within the society's research portal. Influenced by these results, SPE leadership committed to deployment of ATHENA to support improved information retrieval within its portal.

This paper describes advances since that first evaluation aimed at positioning ATHENA for successful deployment to the SPE. These advances include chat-based multi-document retrieval, improved support for answer validation, and more focused proactive dissemination. We present results from a human subjects evaluation showing how these enhancements supported knowledge-intensive tasks related to well planning. ATHENA has been integrated into the SPE's research portal and has been deployed to an initial set of test users; it is planned to be deployed to the broader membership in Q4 of 2025.

## Initial ATHENA Capability and Evaluation

### Initial ATHENA Capability

The initial ATHENA capability (Figure 1), described in detail in (Eckroth et al. 2025), was formulated as a chat-based framework in which users interact with ATHENA to receive answers to questions while they complete tasks.

The chat capability employs an agentic RAG approach (Acharya et al. 2025) in which individual agents target specialized parts of the user's request, performing search for relevant documents and/or chunks that address that aspect of the request (Boden et al. 2025). The top matches are sent to an LLM along with the user's request to generate the response, using chain of thought prompting (Wei et al. 2024) to provide references and rationale.

While performing tasks, users can add insights that may benefit others in their CoP by simply typing text in a co-located capture window. ATHENA augments their inputs with relevant contextual information (information about the current user and task) to facilitate downstream dissemination to other CoP members when working on similar tasks. This design makes it easy for users to contribute knowledge "in the flow" with minimal disruption to current activities. It also avoids the need to learn syntax and nomenclature for manually tagging material or mastery of complex tools.

ATHENA contains more than 2000 insights for potential dissemination. Some were captured from the domain experts who helped formulate our evaluations, some were generated by LLMs analyzing corpus content (e.g., where to find certain types of information, defining acronyms), and some were created by asking ChatGPT to provide insights for specified tasks, drawing on general petroleum engineering knowledge. The quality and relevance of the content in this store was highly variable, introducing challenges for effective dissemination. To balance the desire to find helpful insights with concerns about information overload, ATHENA selected three insights to present for each new task performed by a user. This was done by first selecting 15 insights via an embedding mechanism fine-tuned for tasks within our domain, followed by selecting the final 3 insights using an LLM. Insights were then personalized to users by identifying their expertise on topics relevant to the Oil & Gas domain. Novices and intermediate-level users were provided additional topic-specific definitions.

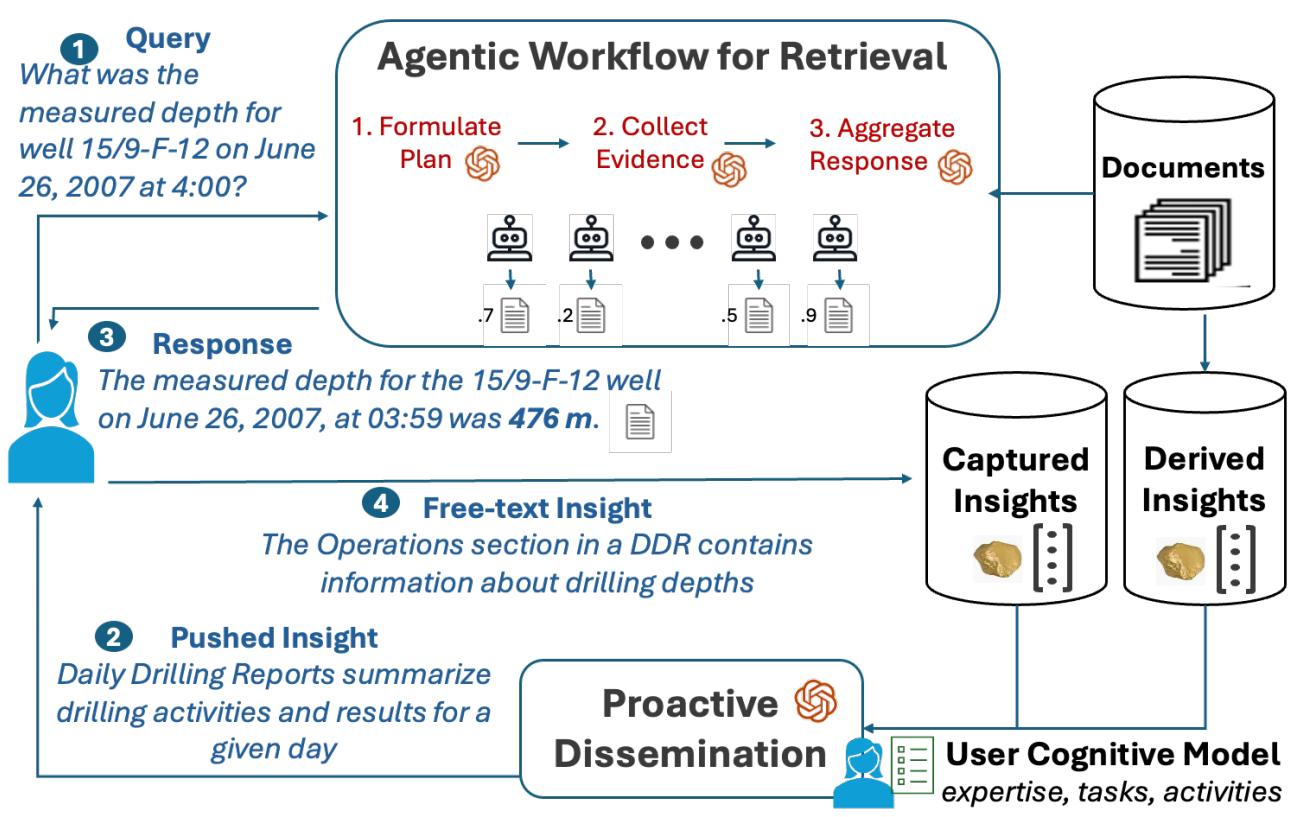

Figure 1. ATHENA overview

### Technology gaps/changes for SPE deployment

ATHENA was received enthusiastically by members of the SPE in our initial evaluation with 79% of them recommending adoption to power the society's research portal. The evaluation also revealed several areas for improvement to better satisfy member needs, which we discuss below.

**Accuracy and Verifiability of Responses.** Safety is a top concern for the Oil & Gas sector, making accuracy of responses critical. An SME with whom we worked noted that petroleum engineers will need to validate all information produced by a chat system, thus elevating the importance of providing rationale for a generated response and the means to quickly verify responses in relevant source material.

**Multi-source aggregation.** The initial ATHENA system delivered answers to questions about both general petroleum engineering knowledge (e.g., details of equations for measuring pore pressure) or information related to drilling activities (e.g., the type of pressure test performed in a given well on a given day, the maximum depth to which a well was drilled). Getting fast, reliable answers to these types of questions greatly facilitates well planning. However, there is also a need to support aggregation of information that may be dispersed throughout a corpus. Examples include compiling the set of lost circulation events for a given well, or summa-rizing candidate offset wells to be used in selecting the location for a new well to be drilled. ATHENA's initial agentic approach to finding information was not capable of collating and extracting knowledge spread across 10's to 100's of documents when answering a question.

**Improved proactive dissemination.** ATHENA's proactive dissemination was generally perceived as helpful in our initial evaluation but there was clear room for improvement. Our top-3 embeddings-based retrieval reliably found at least one useful insight for a task but often included an insight that was not particularly relevant, causing some users to discount their overall utility. Also, there was often overlap in content in the selected insights. While such overlap could build confidence in pushed content, it also required more effort by the user to process. Finally, some of the captured insights were on topic but not clear or actionable, thus limiting their value for completing tasks.

## Technical Advances

To better position ATHENA for deployment, we extended its capabilities for information retrieval, answer verification, and community-authored insight dissemination.

### Multi-Hypothesis Search

ATHENA is proficient at finding the correct documents for answering user questions. Typical RAG approaches include BM25 keyword search and/or finding the top-k closest vector embeddings according to cosine similarity. Due to the complexity of the Oil & Gas domain, specific syntax of well names (such as 15/9 F-1-C), and significance of dates, times, and locations, we found that we needed to utilize additional information beyond keywords and embeddings. ATHENA's updated search works as follows. First, the user's question is rewritten in multiple ways (typically five) to pose different interpretations or hypotheses about the most effective phrasing to search for the relevant documents. Then these interpretations are classified (tagged) according to location including Oil & Gas geography (basins, fields, wells, etc.), a topic (e.g., "adverse event" like lost circulation), and a record type (e.g., daily drilling report, final well report, well log). These tags are then considered as possible filters by an LLM, which is instructed to choose among them and choose keywords for BM25 search.

We benchmarked ATHENA's search method with embedding search according to mean reciprocal rank for retrieving documents that were relevant for the tasks used in the well planning evaluation described in the next section. As the results in Table 1 show, that ATHENA's search improves dramatically on vector embeddings.

As mentioned, typical RAG applications use solely embedding-based similarity retrieval and/or keyword BM25 matching. We experimented with ATHENA's multi-hypothesis search technique using ablation studies. By eliminating some of ATHENA's features, such as forming multiple interpretations of the user's question and eliminating the use of classification as filters, we found that ATHENA's full features achieved 0.90 accuracy at finding the right documents vs. 0.59 accuracy when all ablations were active (replicating keyword BM25 matching).

*Table 1. Mean reciprocal rank (MRR) of ATHENA search versus common vector embedding models.*

| Search Technique | MRR |
|---|---|
| ATHENA | 0.687 |
| MiniLM-L6-v2 (Reimers & Gurevych 2019) | 0.412 |
| text-embedding-3-small (OpenAI 2025) | 0.376 |

### Support for Answer Verification

Documents in the Oil & Gas industry are sometimes 100+ pages. While it is essential to cite sources for LLM-generated answers, citations to whole documents are not useful. We instrumented our document-reading agents with the ability to record the page number and a short quote of the text for any knowledge that is extracted from that document and cited in the answer to the user's question. We also built an integrated document viewer so that the user may click the citation in ATHENA's generated answer, which pops open the document viewer, and then they may jump to the location in the document where the knowledge that was cited originated. This facility works well even for long tables that are commonly found in these documents.

### Agentic Multi-Document Retrieval

To support information aggregation tasks, we introduced a new "batch document reading" agent that iterates (in parallel batches for speed) over all documents or chunks produced by search agents with the goal of extracting specific content from each item. The agent orchestrator (GPT-4o) determines when this batch tool is required based on the user's question. For example, if a user asks for all lost circulation events that occurred in a given well in a specific year, potentially hundreds of daily drilling reports are relevant and must be examined. The orchestrator tasks the batch tool with finding specific knowledge (e.g., "lost circulation events") from each item. The results are then collected and examined one final time by an LLM to make an aggregate response. Raw results are also delivered as a CSV file for inspection.

ATHENA's multi-document retrieval introduces a significant and crucial functionality that is not available in a typical RAG application, which can only review the top-*k* (e.g., top 10) chunks retrieved from the corpus for a given query.

Likewise, even OpenAI's Deep Research and similar tools will not exhaustively examine all sources. The agentic orchestrator in Deep Research appears to stop early when it believes it has found "enough" information to answer the question, but for some questions, only an exhaustive examination of the material can complete the task.

## Proactive Dissemination of Insights

As described above, ATHENA proactively disseminates task-related *insights* tailored to an individual's inferred expertise and problem-solving context, seeking to identify relevant content from a store of more than 2,000 elements captured from humans or generated with the help of LLMs.

Our approach to personalization employs a User Cognitive Model (UCM) that stores information available about the user, specifically: *background* data, e.g., collected via surveys, or from a resumé or organizational records; *interaction* data captured by ATHENA's UI instrumentation, comprising the user's chat history, consulted documents, etc.; and *contextual* data, including a description of the task being performed. The original ATHENA system relied on a short demographics survey to identify a user's expertise on a (small) set of Oil & Gas topics, leading the LLM used for personalization to hallucinate and guess about the user's expertise on topics not in the survey. To improve on that, we extended the UCM by maintaining a user *competency vector* consisting of inferred expertise values on topics. We use competency vectors to identify knowledge gaps and guide the delivery of content to users, resulting in fine-grained modeling of expertise that goes beyond a simple global classification as being a novice or expert in the field.

To form the vectors, we leverage existing taxonomies already used by ATHENA to classify documents in our corpus and search for relevant information during retrieval. Each entry in the vector corresponds to a *topic* node in a taxonomy, resulting in competency vectors with more than 200,000 entries in total. Since background information about a user might be scarce, a direct identification of topics will result in sparse and incomplete competency vectors. However, if we know that a user is, e.g., a drilling engineer, we can infer that he/she is highly familiar with drilling operations, events, reports, etc.

To perform such inference, we implemented a *competency map*, using an LLM (gpt-4o) to directly create associations between taxonomy nodes, and to derive associations between work areas, job titles and skills from an established career guide (EnergyAPI 2018), linking them to the corresponding topic nodes in the taxonomies. To initialize a competency vector for a new user, we identify relevant topics for which we have evidence of the user's expertise and assign numeric values to those and related topics proportional to their association strength, as dictated by the competency map. All values are then normalized to allow specifying thresholds useful at personalization time.

Our insight dissemination pipeline parallels the one used in the initial ATHENA in that it includes a *selection* phase

Insights:

- Baker Hughes' PowerPulse tools are known for their high data transmission rates, utilizing advanced mud pulse telemetry techniques.
- These downhole monitoring devices measure pressure and temperature, providing real-time data that optimizes production and enhances reservoir management.
- PowerPulse tools are often turbine-powered, generating their own power from the flow of drilling mud for continuous operation.
- This capability allows for real-time monitoring of downhole conditions and tool performance, improving decision-making and operational efficiency in oil and gas exploration.
- PowerPulse tools enhance real-time data transmission and monitoring, providing critical insights into geological formations and wellbore conditions to optimize drilling performance.

Hide details

This information was selected since you are a novice in PowerPulse tools and mud pulse telemetry, which are key aspects of the content provided.

- PowerPulse tools are renowned for their high data transmission rates, achieving up to 16 bits per second using advanced mud pulse telemetry techniques like zero-gap modulation and data compression. This is significantly faster than older mud pulse systems.

  *This insight was added by: Sarina Goodman*

- Baker Hughes' PowerPulse tool is comparable to Schlumberger's EcoScope tool.

  *This insight was added by: Prueba Vara*

- Unlike some MWD/LWD tools that rely solely on batteries, PowerPulse tools are often turbine-powered. This means they generate their own power from the flow of drilling mud, allowing for continuous operation without needing to trip out for battery changes.

  *This insight was added by: Yonos Pirathiba*

*Figure 2. Sample disseminated insights*

followed by *personalization*. However, it uses a new selection mechanism that no longer requires a final LLM-based selection step, and LLM-based aggregation and personalization relying on the competency vectors. Overall, these changes address the issues detected during the initial evaluation of ATHENA where insights often provided irrelevant and redundant information.

For selection, whenever a user starts a new task, ATHENA first considers whether any insights in its store are relevant for dissemination given the task's description. We implemented a method that computes a matching score (semantic similarity) between insights and tasks using the cosine distance between embeddings produced by both a function that was trained to closely match the machine-generated insights with related tasks in latent space using a contrastive loss, and a TF-IDF approach (Spärck Jones, K. 1972) that identifies relevant topics in the insight and task descriptions. (The use of TF-IDF was motivated by a limitation identified while training the ML embedding function using machine-generated insights, which are out-of-distribution w.r.t. human-captured ones, leading to low rankings for insights that are critical for the evaluation tasks.) The weighted average between the two distances (20% for the ML embedding, 80% for TF-IDF) is used to rank the insights in order of relevance for the task. Compared to the previous approach relying solely on embeddings, the new approach achieves a 500% MRR (inverse rank) improvement and 84% decrease on the average number of insights ($k$) needed to ensure the selection of critical insights associated with our evaluation tasks.

The top-$k$ selected insights are then tailored to the user. Similar to our previous approach, we first use an LLM to identify (offline) key topics given each insight's description, map them to ATHENA's taxonomies, and generate definitions targeted at novice- and intermediate-level users. However, at insight personalization time (online, following insight selection), we now use the user's competency vector to classify his/her topic-specific expertise as either novice, intermediate, or expert for each of the key topics associated with the top-$k$ insights using predefined thresholds. The precomputed definitions are appended to the insights, and all information is then fed to an LLM which is tasked to combine them into a single, coherent text snippet to help the user complete the task.

Next, given the task description, the user's competency vector (subset of relevant topics) and the combined insight, another LLM call is made to remove redundant and irrelevant information, keeping only task-specific, essential information matching the user's level of expertise. For example, insights related to topics for which the user is considered an expert are removed or only briefly mentioned, whereas information related to topics for which the user is deemed a novice is kept intact from the original insight and expanded using the precomputed concept definitions. The final combined, personalized, and sanitized insight is then decomposed into text bullets for delivery to the user.

Figure 2 shows an example of insights delivered as a response to a user request: "*Make a table of the planned MWD measurements taken with PowerPulse for well 15/9-F-12, including the PowerPulse tool and the section in which it was planned to be used.*" The user is overall an expert in well planning but works for a company that uses an alternative to PowerPulse and so is not familiar with it. ATHENA's UI presents the information bullets resulting from insight selection and personalization below the user's query, in a separate box. The user can click a "Show details" button to robtain rationale for the pushed insights, which is generated by an LLM given the information derived during personalization (see below "Hide details" in Figure 2). A list of the original insights and their provenance is also provided. This design facilitates verification of information provided by the chat interface, increasing user trust in ATHENA.

## Evaluation

### Experimental Protocol

We ran a within-subjects experiment involving 13 subjects with technical backgrounds but no expertise in petroleum engineering to evaluate our updated ATHENA system. While we would have liked to have used SPE members again, the society felt it was unnecessary given their intent to proceed with deployment. Given our prior finding that ATHENA enabled non-experts to perform at levels comparable to SPE members, we felt confident that our testing with non-SPE subjects would readily transfer to SPE members.

Subjects performed two complexity-matched well-planning tasks with each of ATHENA and a baseline RAG system using the same corpus, with system order counter-balanced across subjects to avoid order effects and tasks assigned randomly. Within each task pair, we reused one from the earlier evaluation and one new task that focused on aggregating information from multiple sources. For proactive dissemination, we used top-$k = 3$ for insight selection.

The baseline was built by a third party tasked by our research funder to create a state-of-the-art tool for comparison. It combined page-level chunking indexed via embeddings (Cohere v3) for check retrieval with a keyword-based index for extracting dates and well names. Claude 3.5 Haiku was used to map a given query to the appropriate index to use (possibly); Claude 3.7 was then used to generate the response along with links to chunks used in the response.

### Results

The results in Table 2 show that ATHENA enabled superior task performance compared to the baseline across all measures. Average task grade improved by 216% while subject failures were eliminated. Follow-up t-tests showed the improvement in task grade is significant ($p < .0001$).

Task timing results proved problematic as many subjects reported simply giving up in the baseline condition because the tool was not helpful in identifying needed materials. We also saw many baseline responses indicating that subjects were entering arbitrary responses so they could proceed to the next task. For this reason, we report *productivity* (task grade scaled by minute spent) rather than task durations. As can be seen from the table, productivity tripled.

Table 2 also reports scores for a novel Knowledge Management System Usability Scale (KM-SUS; Keefe et al., 2025). The KM-SUS was developed to evaluate the usability of KM systems, considering the unique tasks, expectations, and goals that users have as they interact with KM systems. As shown, subjects rated ATHENA significantly more usable than the Baseline ($p < .05$).

Additional feedback showed subjects found the disseminated insights to be helpful for finding relevant material, for validating chat responses, and for explaining terminology; several noted that they also found the insights to be educational. In response to the question *"Were the pushed insights helpful?"* 82% of the 11 respondents found the insights to be valuable for completing the evaluation tasks.

We manually identified some SME-curated insights as being a *priority* for each evaluation task, meaning they are critical for task performance, and *desirable insights* to be provided that are not critical. ATHENA attained perfect recall for priority insights and $0.90 \pm 0.30$ for desirable ones, demonstrating that all relevant information was provided to users at the *right time*. Moreover, an average of $2.48 \pm 1.13$ insights was selected for dissemination, resulting in $2.22 \pm 1.30$ information bullets presented after aggregation and personalization. The dissemination mechanism condensed insights by $\sim$18%, as measured by the ratio between the total length of delivered text bullets and the selected original insights (mean ratio $0.82 \pm 0.34$), showing that the new aggregation and personalization mechanism removes redundant and irrelevant data while retaining crucial information.

*Table 2. Evaluation Results (N = 13)*

| | Avg. Grade [0-100] | Failure Rate (grade<50%) | Productivity (grade/minute) | KM-SUS |
|---|---|---|---|---|
| Baseline | 27.2 | 62% | 3.2 | 32.3 |
| ATHENA | 85.8 | 0% | 10.2 | 41.2 |
| Change | 216% | -100% | 215% | 28% |

## Path to Deployment

Starting in 2015, the SPE and i2k Connect worked together to develop the SPE Research Portal to address the challenges facing engineers who are looking for technical information (Smith, R.G. et al. 2021) (https://search.spe.org/). The portal uses knowledge-based faceted classification, entity recognition, title discovery, summarization and concept tag identification for finding, analyzing, and visualizing SPE content. The portal currently contains 311,322 items from SPE and 22 related technical societies.

The initial focus for the deployment is chat-based retrieval, which we have integrated with other portal capabilities so users can take advantage of maps, filtering, and other analysis and visualization tools. Longer term, SPE is interested in potentially deploying ATHENA's capture and dissemination capabilities to support information sharing throughout the CoPs within its membership.

The initial system was deployed in June 2025 to 25 early adopters engaged in an SPE project to identify best practices papers in Well Completion and to create new synthesis papers. SPE has approved broader roll-out for Q4 2025.

One issue that had to be addressed prior to SPE sanctioning deployment is the business case. The new LLM-based technology comes with a significant uplift in cost compared to the current SPE Research Portal. We worked with SPE to develop a model that balances the increased cost with the increased member benefits that generative AI brings.

Our success in moving our research to deployment was greatly facilitated by the following non-technical factors.

- The team includes several members with deep industry and knowledge management experience, in addition to their AI experience.
- The team has built credibility with the SPE over the course of a 10-year relationship. One team member was formerly the executive manager at the SPE responsible for starting the relationship; his credibility within SPE and the broader industry community has been invaluable.
- While developing ATHENA, more than 144 SPE members participated in evaluations of the evolving technology, helping to keep our work aligned with SPE needs.
- We meet with SPE staff on a weekly basis to address issues and to identify improvement opportunities. We also engage regularly with SPE leadership.

## Conclusions

We describe advances to our virtual member of a Community of Practice for the Oil & Gas community, called ATHENA, designed to position it for successful deployment to support chat-based information retrieval for the SPE Research Portal. Specifically, we expanded an initial high-accuracy question-answering capability to a more broadly scoped agentic retrieval method that can further assist with tasks that require gathering information from across numerous documents within a corpus. We also redesigned the original proactive information dissemination capability to increase the relevance of selected content and to better tailor it to a given context and user. An evaluation involving realistic well-planning tasks echoed earlier findings that ATHENA enables significant productivity gains compared to a state-of-the-art RAG baseline.

ATHENA was deployed in June 2025 for use by 25 early adopters at the SPE; broader roll-out the SPE membership is planned for Q4 2025. Going forward, we will work with SPE to understand how the capability is being used in a way that preserves individual and organizational privacy. We will also continue to engage in regular, formal evaluations to assess the utility of the technology for SPE members.